\documentclass{article}

\usepackage{microtype}
\usepackage{graphicx}
\usepackage{float}
\usepackage{array}
\usepackage{subfigure}
\usepackage{amsmath}
\usepackage{amssymb}
\usepackage{svg}
\usepackage{algpseudocode}
\usepackage{booktabs} % for professional tables
\usepackage{xspace}

\newcommand{\ghomala}{Ghom\'{a}l\'{a}\xspace}
\newcommand{\ewe}{\'{E}w\'{e}\xspace}

\usepackage{hyperref}

\usepackage[accepted]{icml2018_ift6269}
\begin{document}

\twocolumn[
\icmltitle{Beyond MLE: Investigating SEARNN for Low-Resourced Neural Machine Translation}



\icmlsetsymbol{equal}{*}

\begin{icmlauthorlist}
\icmlauthor{Chris Emezue}{to}
% \icmlauthor{Bauiu C.~Yyyy}{equal,to,goo}
% \icmlauthor{Cieua Vvvvv}{goo}
% \icmlauthor{Iaesut Saoeu}{ed}
% \icmlauthor{Fiuea Rrrr}{to}
% \icmlauthor{Tateu H.~Yasehe}{ed,to,goo}
% \icmlauthor{Aaoeu Iasoh}{goo}
% \icmlauthor{Buiui Eueu}{ed}
% \icmlauthor{Aeuia Zzzz}{ed}
% \icmlauthor{Bieea C.~Yyyy}{to,goo}
% \icmlauthor{Teoau Xxxx}{ed}
% \icmlauthor{Eee Pppp}{ed}
\end{icmlauthorlist}

\icmlaffiliation{to}{Mila, Lanfrica}
% \icmlaffiliation{goo}{Googol ShallowMind, New London, Michigan, USA}
% \icmlaffiliation{ed}{School of Computation, University of Edenborrow, Edenborrow, United Kingdom}

\icmlcorrespondingauthor{Chris Emezue}{chris.emezue@mila.quebec}
%\icmlkeywords{Machine Learning, ICML}

\vskip 0.3in
]



\printAffiliationsAndNotice{}  % leave blank if no need to mention equal contribution

\begin{abstract}
Structured prediction tasks, like machine translation, involve learning functions that map structured inputs to structured outputs. Recurrent Neural Networks (RNNs) have historically been a popular choice for such tasks, including in natural language processing (NLP) applications. However, training RNNs using Maximum Likelihood Estimation (MLE) has its limitations, including exposure bias and a mismatch between training and testing metrics. SEARNN, based on the learning to search (L2S) framework, has been proposed as an alternative to MLE for RNN training. This project explored the potential of SEARNN to improve machine translation for low-resourced African languages -- a challenging task characterized by limited training data availability and the morphological complexity of the languages. Through experiments conducted on translation for English to Igbo, French to \ewe, and French to \ghomala directions, this project evaluated the efficacy of SEARNN over MLE in addressing the unique challenges posed by these languages. With an average BLEU score improvement of $5.4$\% over the MLE objective, we proved that SEARNN is indeed a viable algorithm to effectively train RNNs on machine translation for low-resourced languages.

\end{abstract}

\begin{figure*}[!h]
    \centering
    \resizebox{1.7\columnwidth}{!}{\includegraphics[]{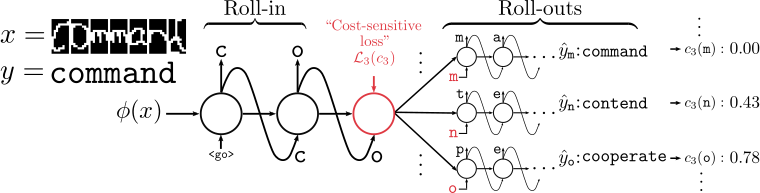}}
    \caption{``Illustration of the roll-in/roll-out mechanism used in SEARNN. The goal is to obtain a vector of costs for each cell of the RNN in order to define a cost-sensitive loss to train the network. These vectors have one entry per possible token. Here, we show how to obtain the vector of costs for the red cell. First, we use a roll-in policy to predict until the cell of interest. We highlight here the learned policy where the network passes its own prediction to the next cell. Second, we proceed to the roll-out phase. We feed every possible token (illustrated by the red letters) to the next cell and let the model predict the full sequence. For each token $a$, we obtain a predicted sequence $\hat{y}_a$. Comparing it to the ground truth sequence $y$ yields the associated cost $c(a)$." -- \citet{searnn}
}
    \label{fig:searn-drawing}
    
\end{figure*}
\section{Introduction}

Prediction is the task of learning a function $f$ that maps inputs $x$ from an input domain $\mathbb{X}$ to outputs $y$ in an output domain $\mathbb{Y}$ \cite{searn}. In structured prediction, the inputs and/or outputs have a structure. For example, in machine translation (or generally speaking any sequence-to-sequence task), the inputs and outputs are sentences: a fundamental structure of any sentence $s$ is that it decomposes into dependent sub-components $s = (s_{1},...,s_{T})$. Effectively addressing structured prediction tasks inherently involves utilizing algorithms that harness both the local and global information embedded within their structure.%Structured prediction, especially sequence-to-sequence tasks, is immensely useful and has a large variety of real world applications.

To tackle sequence-to-sequence tasks, a traditional type of neural networks called recurrent neural networks (RNNs for short) were the go-to solution for a period, until they were gradually superseded by Transformers \cite{vaswani2017attention}. RNNs have proven useful in tackling structured prediction tasks like machine translation \cite{sutskever2014sequence}, parsing, as well as many other tasks in natural language processing (NLP). Beyond NLP, they have been applied to domains like computer vision \cite{sinha2018deep} and reinforcement learning \cite{li23twostage}. The standard RNNs have a recurrent cell or unit to generate tokens sequentially, with each prediction dependent on the preceding ones, allowing the cell to model the joint distribution of the sequence as a product of conditional distributions \cite{searnn}. 

The standard loss used for training RNNs (and to a large extent most sequence-to-sequence prediction tasks) is the maximum likelihood estimation (or MLE loss for short). During RNN training, we operate under the assumption that at each step of the sequence, the cell produces a probability distribution over the tokens, and therefore aim to maximize the likelihood of the ground truth token. 

While the MLE loss is widely used and has been very effective, it has some limitations in sequence-to-sequence learning. According to \citet{searnn}, one issue is called \textit{exposure bias}: during training, the model learns based on the correct answers, but during testing, it has to rely on its own guesses. This can lead to under-performance during deployment as the model might not have learned how to recover from its mistakes. Another limitation is that the training method does not match well with the different ways we measure success in testing or deployment. For example, during testing we might be interested in BLEU score for machine translation not MLE. %For example, it does not consider how close a prediction is to the right answer, only if it's right or wrong. This means it does not use all the information available to improve its predictions.

\cite{searnn} proposed the SEARNN algorithm, an alternative method to train RNNs on sequence-to-sequence prediction -- one that combines searching and learning, builds on the SEARN framework \cite{searn}, and mitigates the exposure bias. SEARNN is based on the learning to search (L2S) approach to structured prediction. The central idea of L2S is to reduce the structured prediction problem to a problem of cost-sensitive classification learning \cite{searn}.

%Personally, I am a fan of the L2S framework.
The L2S framework serves as a source of inspiration for me due because from my point of view, it smoothly blends exploration and exploitation, akin to reinforcement learning (RL). Furthermore, due to its multi-step future in-look and the inability to incorporate useful constraints in the cost, I perceive the algorithm as potentially beneficial, particularly in scenarios characterized by limited training data (which means a standard model trained on MLE is unable to explore diverse configurations of the language). The scant availability of data for (as well as the morphological complexity of) African languages often results in subpar performance in natural language processing (NLP) tasks \cite{masakhanePaper,adelani2022thousand,emezue2022mmtafrica}. I posit that frameworks such as SEARNN hold promise in such contexts of data scarcity and complex tasks. This project therefore aims to investigate this conjecture by employing the SEARNN pipeline to train an RNN on machine translation in three directions involving three interesting African languages: English $\rightarrow$ Igbo (en$\rightarrow$ibo), French $\rightarrow$ \ewe (fr$\rightarrow$ewe), and French $\rightarrow$ \ghomala (fr$\rightarrow$bbj).

\section{The SEARNN algorithm}
The SEARNN algortihm was proposed by \citet{searnn}. Below, I recapitulate an aspect of the SEARNN algorithm written by \citet{searnn}.

``\textit{The basic idea of the SEARNN algorithm is quite simple: we borrow from L2S the idea of using a \emph{global} loss for each \emph{local} cell of the RNN. As in L2S, we first compute a \emph{roll-in} trajectory, following a specific roll-in policy. Then, at each step $t$ of this trajectory, we compute the costs $c_t(a)$ associated with each possible token $a$. To do so we pick $a$ at this step and then follow a \emph{roll-out} policy to finish the output sequence $\hat{y}_a$. We then compare $\hat{y}_a$ with the ground truth using the test error itself, rather than a surrogate. By repeating this for the $T$ steps we obtain $T$ cost vectors. We use this information to derive one \emph{cost-sensitive} training loss for each cell, which allows us to compute an update for the parameters of the model.}" -- \citet{searnn}

Figure \ref{fig:searn-drawing} by the same authors describes the mechanism (using the task of OCR recognition) by which SEARNN obtains the cost-sensitive loss to train the network.

% The objective is to derive a cost vector for each cell of the RNN, which is crucial for defining a loss that is sensitive to costs and enables effective training of the network. These vectors comprise entries corresponding to each possible token. Here, we elucidate the process of obtaining the cost vector for the red cell. Initially, a roll-in policy is employed to predict until reaching the cell of interest. Notably, this entails utilizing a learned policy where the network feeds its own prediction to the subsequent cell. Subsequently, the roll-out phase ensues. All possible tokens (depicted by the red letters) are input to the subsequent cell, allowing the model to predict the entire sequence. For each token \(a\), a predicted sequence \(\hat{y}_a\) is obtained, and its comparison with the ground truth sequence \(y\) yields the corresponding cost \(c(a)\).

% First, the current model is used to make a new set of examples to train a classifier (in this case, it is another RNN). These new training examples have a special feature: each one has a list of `costs', one for every possible word. To get these costs, the model guesses what all the words will be up to a certain point. Then, at each step, it tries out different guesses for what the next word might be and keeps guessing until it finishes a sample sentence (determined by reaching an apriori given word length). After that, it figures out how much each sentence would cost.

\paragraph{The roll-in and roll-out policies:}
The roll-in policy determines how to obtain the context while the roll out policy generates target completions that allow us to calculate the cost between the predicted target completions and the ground-truth. Intuitively, the roll-in policy can be understood as dictating the portion of the search space explored by the algorithm, while the roll-out policy is looking into the future to observe various possible endings for the sequence and their associated costs. This "looking into the future" approach is significantly different from MLE training and offers key advantages: for one, it allows the model to learn to search through the space (similar to exploration in RL) which, in theory, gives the system more exposure, thereby making its predictions more robust, and diverse. There are various options for the roll-in and roll-out policy, all of which have been extensively researched by \citet{chang2015learning}. Using a `reference' policy corresponds to using the ground-truth tokens. For example, the `reference' roll-in policy is similar to teacher-forcing in seq2seq learning, where the conditional probability of the next token is based on the ground-truth tokens. A `learned' policy on the other hand corresponds to using the learned model itself to estimate the conditional probability. Finally, it is possible to leverage a `mixed' approach that (perhaps stochastically) combines both \cite{chang2015learning,searnn}.

\paragraph{Choosing a cost-sensitive loss:}
\citet{searnn} proposed two cost-losses for SEARNN training: the log-loss (LL), and the Kullback-Leibler divergence loss (KL). The LL loss is structurally similar to MLE, with the main difference being that instead of maximizing the probability of
the ground truth action, we maximize the probability of the best performing action at the current step $t$ with respect to the cost vector $c_{t}$ \cite{searnn}: 

\begin{equation}
	\mathcal{L}_t(s_t; c_t) = -\log\Big(e^{s_t(a^*)} \text{\big/} \sum_{i=1}^A e^{s_t(i)}\Big),
	\end{equation}
 where $a^*=argmin_{a \in \mathcal{A}} c_t(a)$.
The KL loss transforms the LL loss into a probability matching loss by transforming each cost vector into a probability distribution (e.g. through a softmax operator) and then minimizing a divergence between the current model distribution $P_M$ and the ``target distribution'' $P_C$ derived from the costs. Since the costs are considered fixed with respect to the parameters of the model, our loss is equivalent to the cross-entropy between $P_C$ and $P_M$ \cite{searnn}.

\begin{equation}
\mathcal{L}_t(s_t; c_t) = - \sum_{a=1}^A \Big(P_C(a) \log\big(P_M(a)\big)\Big) \quad
\end{equation}
\begin{equation*}
\begin{aligned} &\text{where } \; P_C(a) = e^{-\alpha c_t(a)} \text{\big/} \textstyle\sum_{i=1}^A e^{-\alpha c_t(i)} \\
& \text{and} \; P_M(a) = e^{s_t(a)}\text{\big/} \textstyle\sum_{i=1}^A e^{s_t(i)}.
\end{aligned}
\end{equation*}

$\alpha$ is a scaling parameter that controls how peaky the target distributions are. It can be chosen using a validation set.

\subsection{Scaling SEARNN for machine translation}
\label{sec:scaling}
The SEARNN algorithm is computationally expensive  as it performs a large number of roll-outs. This computation cost grows with the length of the sequence and the number of possible tokens (i.e. the vocabulary size). This creates a bottleneck when the algorithm is applied to tasks with large vocabulary sizes and output sequences, a typical scenario in machine translation \cite{searnn}. To mitigate this bottleneck, \citet{searnn} introduced a variety of sampling techniques. They further adapted the original cost-loss to be compatible with sampling on the tokens. Finally, they showed that the sampling variants of SEARNN still outperform MLE, while attaining a $5X$ running time speedup. For the experiments in this project, sampling was employed according to the adopted settings of \citet{searnn}.

\section{Experiments}
My application project involved applying the SEARNN approach to train an RNN on machine translation for English $\rightarrow$ Igbo (en$\rightarrow$ibo), French $\rightarrow$ \ewe (fr$\rightarrow$ewe), and French $\rightarrow$ \ghomala (fr$\rightarrow$bbj). The aim of this project was to explore whether there will be performance improvement with using the SEARNN algorithm to train the RNN over the standard MLE-based RNN training. For the baseline, I trained the same RNN architecture using the MLE objective.

\subsection{Languages considered}
I chose \'{E}w\'{e} and Ghom\'{a}l\'{a} for three reasons: 1) they have received very rare attention in machine translation for African languages \cite{adelani2022thousand}; 2) these languages have a degree of morphological complexity and alphabetical structure that is very different from English and French languages \cite{Noss_1984}; 3) they have very few training samples. I chose Igbo language because it is my native language, and in this way I could observe the translations from both training algorithms to evaluate their quality.

\subsection{Data}
I leveraged the MAFAND-MT Corpus \cite{adelani2022thousand} -- a corpus of texts curated from news sources in the native language and translated by professional translators in their respective source language. The source language chosen depended on whether the country where the language is majorly spoken is predominantly Anglophone or Francophone. Table \ref{tab:data_stat} outlines the details of the corpus for the three languages (note that the original MAFAND-MT Corpus contains more languages). The table also shows the split sizes and we see that \ewe and \ghomala have way smaller samples (and number of speakers) than Igbo, making these languages interesting for this study. I used the train/dev/test split provided by the MAFAND-MT Corpus.

\begin{table*}[h!]
 \footnotesize
 \begin{center}
 \resizebox{\textwidth}{!}{%
    %\scalebox{0.9}{
  %\begin{tabular}{lp{33mm}p{10mm}r|r|p{34mm}r|lr}
  \begin{tabular}{lllr|r|p{52mm}r|lr}
    \toprule
     \textbf{Target}& &\textbf{African} & \textbf{No. of}  & \textbf{Source} & \multicolumn{2}{c|}{\textbf{NEWS}}  & \multicolumn{2}{c}{\textbf{REL}} \\
    \textbf{Language} & \textbf{Family} & \textbf{Region} & \textbf{Speakers}  & \textbf{Lang.} & \textbf{Source} & \textbf{Split Sizes} & \textbf{Source} & \textbf{Total Size} \\
    \midrule
    Ghom\'al\'a' (\texttt{bbj}) & NC / Grassfields &Central& 1M & French & Cameroun Web & 2232/ 1133/ 1430 & Bible & 8K \\
    \'{E}w\'{e} (\texttt{ewe}) & NC / Kwa &West& 7M & French & Benin Web TV & 2026/ 1414/ 1563 & JW300 & 618K \\
    Igbo (\texttt{ibo}) & NC / Volta-Niger &West& 27M & English & \cite{Ezeani2020IgboEnglishMT} & 6998/ 1500/ 1500 & JW300 & 415K \\
    \bottomrule
  \end{tabular}
  }
  \vspace{-3mm}
  \caption{\textbf{Languages and Data Details for MAFAND-MT Corpus}. Language, family (NC: Niger-Congo), number of speakers, news source, news (\texttt{NEWS}), and religious domain (\texttt{REL}) data split.}
  \vspace{-4mm}
  \label{tab:data_stat}
  \end{center}
\end{table*}

\subsection{Model}
\begin{figure}[H]
    \centering
    \includegraphics[width=\columnwidth]{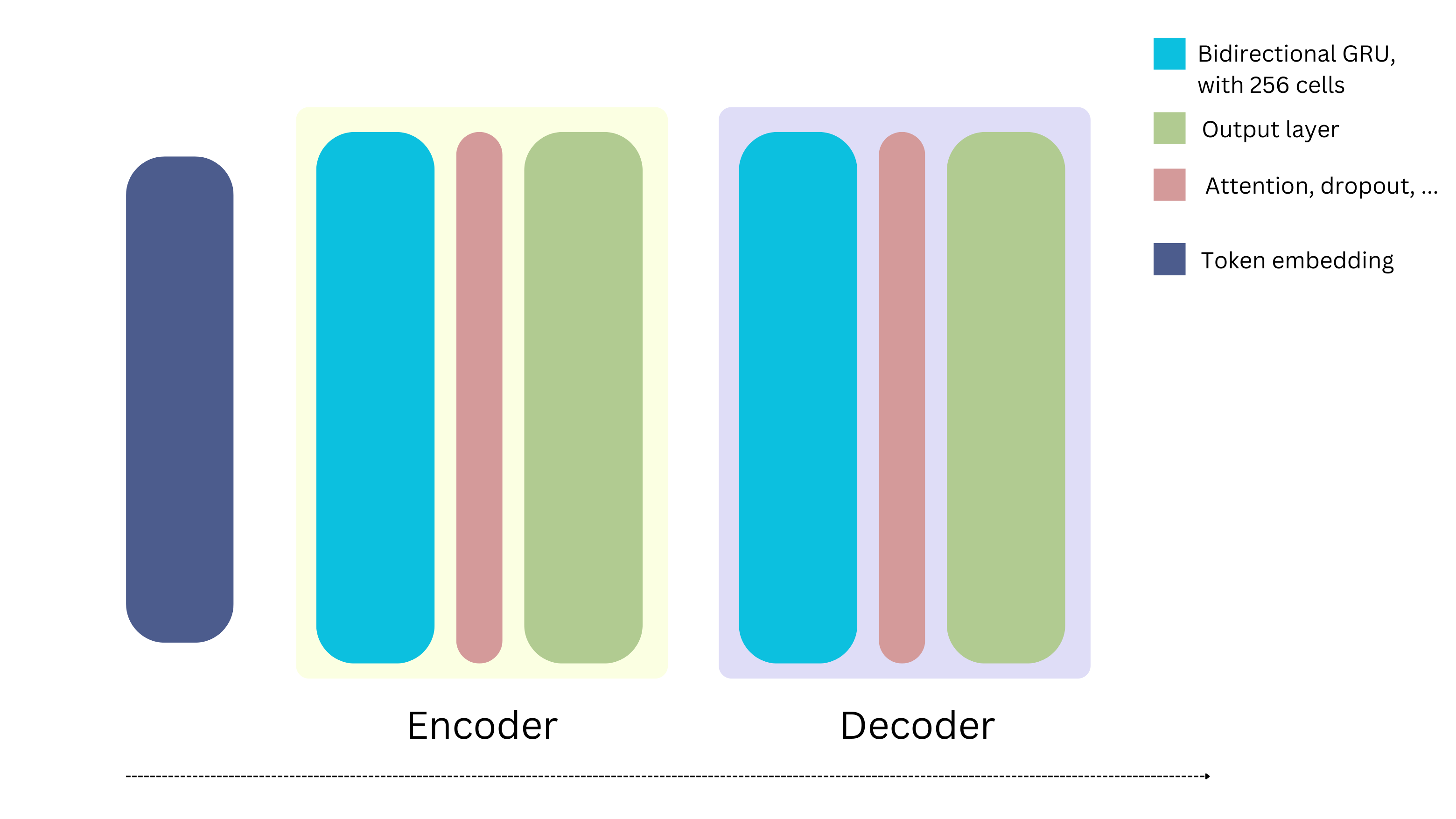}
    \caption{NMT model architecture}
    \label{fig:nmt-architecture}
\end{figure}

Following \citet{searnn}, I used an encoder-decoder model with GRU cells of size 256, with a bidirectional encoder and single-layer RNNs. I illustrated the model architecture in Figure \ref{fig:nmt-architecture}. 

This was a relatively small model compared to contemporary model sizes, so we did not anticipate exceptional results, but rather this experiment served as a proof of concept study.

\subsection{Experimental Settings}
In my SEARNN experimental setup, I used a reference roll-in and a mixed roll-out policy. The KL loss was leveraged in all experiments. For sampling the tokens (Section \ref{sec:scaling}), I employed the custom sampling approach designed by \citet{searnn}, sub-sampling 25 tokens at each cell. The custom technique involved using top-k sampling (where we took the top k tokens according to the current policy) for the first 15 tokens and then sampling 10 neighbouring ground truth labels around the cell. The rationale for these neighboring tokens is that skipping or repeating words is quite a common mistake in NMT \cite{searnn}. To compute the cost, the smoothed BLEU score \cite{bahdanau2016actorcritic} was utilized.
In terms of training settings, I used the Adam optimizer with a learning rate of $1e-3$ and an annealed learning rate scheduler.%Following the authors, I used a tanh attention

\subsection{Workflow}
In this section, I describe the step-by-step implementation process, as well as issues discovered and how I solved them. 

The practical implementation of the SEARNN algorithm to my use case was pretty straightforward. To start with, I forked the authors' repository: \url{https://github.com/chrisemezue/SeaRNN-African}. Thanks to their detailed instructions, I was able to effortlessly navigate through the installation process of the package as well as the corresponding scripts needed to execute the algorithm.

I used the default hyperparameters of the authors' code. The bulk of my time was spent trying to resolve issues due to the old PyTorch and Python versions the authors used (\texttt{torch v0.3}) during the time of running their experiments. Attempting to upgrade the packages led to breakage errors so I had to work with the old versions. The most notable issues were:

\begin{itemize}
    \item there was an issue related to \texttt{torch v0.3} differentiating between \texttt{Variable} and \texttt{tensor} in the \texttt{masked\_fill} function. The latest versions do not do this. Finding the source of this bug was a bit tricky, as online sources either never had this bug (since more updated versions of \texttt{torch} were already being used) or simply proposed upgrading the version of \texttt{torch}. \hyperlink{https://github.com/chrisemezue/SeaRNN-African/commit/8327680cd98292b337bcdb7428466ae0bd5703e0}{My solution} was to transform the mask into a \texttt{Variable}, where it was required, and set \texttt{requires\_grad = False} so no gradients pass through it.

\item Due to perhaps Python 3.5's limited ability to handle very large numbers, the run was constantly throwing a Runtime error with the \texttt{float(-"inf")} used in the same \texttt{masked\_fill} function. After debugging, I resolved to use \texttt{-1e35}: it was the largest number for which I was not receiving the runtime error. I then had to perform unit tests to ensure that I was obtaining an acceptable range of values after the \texttt{masked\_fill} (and subsequent) operations.

\item The SEARNN algorithm was significantly slower than training with MLE objective, even with the sampling technique. I therefore reduced the maximum number of iteration steps from 50,000 to 25,000. Furthermore, I parallelized the experiments across the three translation directions and two training algorithms. The English $\rightarrow$ Igbo direction took about 2-3 days to train, while the rest took 1-2 days.
\end{itemize}

% \begin{itemize}
%     \item Experiments
%     \begin{itemize}
%     \item Formal description of the NMT task
%     \item Formula for how incorporating it with L2S will look like
%     \end{itemize}
%     \item Languages
%     \begin{itemize}
%     \item Why were these languages chosen?
%     \item What makes them interesting to use for this experiment
%     \end{itemize}
%     \item Data
%     \begin{itemize}
%     \item Data details; why this data was chosen
%     \item Split numbers
%     \end{itemize}
%     \item Model
%     \begin{itemize}
%     \item What was the model considered
%     \item Rough overview of the architecture of the model
%     \item Parameters
%     \end{itemize}
%     \item Workflow
%     \begin{itemize}
%     \item What I did in my experiments
%     \item I think here we want to directly pinpoint my step-by-step process of doing this project (up to the experiment part).
%     \end{itemize}
% \end{itemize}

\section{Results \& Discussion}
The corpus-wide BLEU score on the test sets is reported in Table \ref{tab:result}. We observe consistent improvements of the SEARNN algorithm over the MLE objective across all three directions. This corroborates my hypothesis and findings of \citet{searnn} about the superior performance of SEARNN. The BLEU score itself is very low due to the very shallow network architecture used, as well as the small training steps. The findings from Table \ref{tab:result} indicate that there is indeed potential for leveraging the SEARNN algorithm to train RNNs on machine translation for low-resourced languages. 

\begin{table}[H]
\caption{Comparison of SEARNN with MLE on machine translation.}
\label{tab:result}
\vskip 0.15in
\begin{center}
\begin{small}
\begin{sc}
\resizebox{\columnwidth}{!}{
\begin{tabular}{lcccr}
\toprule
 & English $\rightarrow$ Igbo & French $\rightarrow$ \ewe & French $\rightarrow$ \ghomala \\
\midrule
SEARNN   &\textbf{0.0279} & \textbf{0.0018}& \textbf{0.0015}\\
MLE & 0.0276& 0.0010& 0.0010\\
\bottomrule
\end{tabular}
}
\end{sc}
\end{small}
\end{center}
\vskip -0.1in
\end{table}

\begin{figure}[H]
    \centering
    \includegraphics[width=\columnwidth]{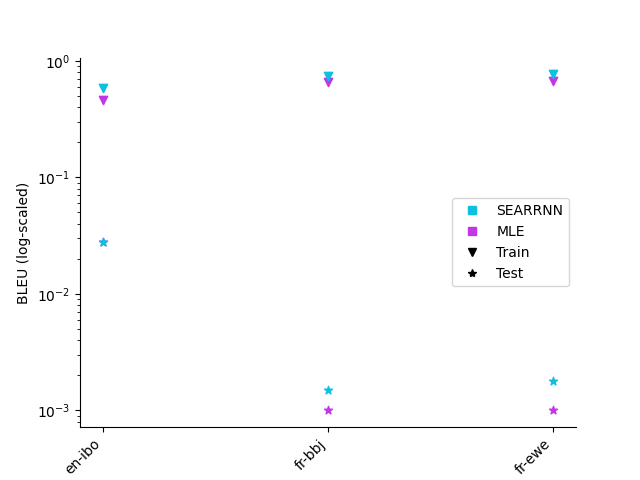}
    \caption{BLEU score on train and test set for the three translation directions. %We observe the superior performance of SEARNN over the MLE objective in both the train and test sets of all translation directions. This shows that even within the training set, the SEARNN algorithm is leading to improvement gains. Furthermore, the test results highlights the better generalization performance of SEARNN
    }
    \label{fig:result}
\end{figure}

To understand the performance on the training outcome, I plotted in Figure \ref{fig:result}, the BLEU scores for both the train and test sets in three different directions. We notice that across all translation directions, SEARNN performs better than the MLE objective in the train sets (which signifies that the training dynamics of SEARNN are better). This indicates that even during training, the SEARNN algorithm leads to better results. Additionally, the test results emphasize that SEARNN has better generalization performance.

\section*{Conclusion}
In this project, I set out to explore the potential of SEARNN in improving machine translation for African languages. Through experiments conducted on translation tasks involving English to Igbo, French to \ewe, and French to \ghomala, I aimed to evaluate the efficacy of SEARNN in addressing the unique challenges posed by these languages, characterized by limited data availability and morphological complexity. Overall, the experimental results suggest that there is indeed potential for leveraging the SEARNN algorithm to train RNNs on machine translation for low-resourced languages. 

There is headroom to explore improvements to SEARNN. For one, figuring out ways to further speed up training time will be helpful. For another, integrating SEARNN with state-of-the-art architectures like the Transformer could lead to improvements in NMT for these languages.
\section*{Acknowledgements}
I acknowledge the substantial contribution of \citet{searnn} as a primary source of the guidance, inspiration and content of this report. Table \ref{tab:data_stat} was obtained from \citet{adelani2022thousand}. Lastly, this research was enabled in part by compute resources provided by Mila (mila.quebec).


\bibliography{example_paper}
\bibliographystyle{icml2018}

% %%%%%%%%%%%%%%%%%%%%%%%%%%%%%%%%%%%%%%%%%%%%%%%%%%%%%%%%%%%%%%%%%%%%%%%%%%%%%%%
% %%%%%%%%%%%%%%%%%%%%%%%%%%%%%%%%%%%%%%%%%%%%%%%%%%%%%%%%%%%%%%%%%%%%%%%%%%%%%%%
% % DELETE THIS PART. DO NOT PLACE CONTENT AFTER THE REFERENCES!
% %%%%%%%%%%%%%%%%%%%%%%%%%%%%%%%%%%%%%%%%%%%%%%%%%%%%%%%%%%%%%%%%%%%%%%%%%%%%%%%
% %%%%%%%%%%%%%%%%%%%%%%%%%%%%%%%%%%%%%%%%%%%%%%%%%%%%%%%%%%%%%%%%%%%%%%%%%%%%%%%
% \appendix
% \section{Do \emph{not} have an appendix here}

% \textbf{\emph{Do not put content after the references.}}
% %
% Put anything that you might normally include after the references in a separate
% supplementary file.

% We recommend that you build supplementary material in a separate document.
% If you must create one PDF and cut it up, please be careful to use a tool that
% does not alter the margins, and that does not aggressively rewrite the PDF file.
% pdftk usually works fine. 

% \textbf{Please do not use Apple's preview to cut off supplementary material.} In
% previous years it has altered margins, and created headaches at the camera-ready
% stage. 
% %%%%%%%%%%%%%%%%%%%%%%%%%%%%%%%%%%%%%%%%%%%%%%%%%%%%%%%%%%%%%%%%%%%%%%%%%%%%%%%
% %%%%%%%%%%%%%%%%%%%%%%%%%%%%%%%%%%%%%%%%%%%%%%%%%%%%%%%%%%%%%%%%%%%%%%%%%%%%%%%

\end{document}